\documentclass[11pt,a4paper]{article}
\usepackage[hyperref]{acl2021}
\usepackage{times}
\usepackage{latexsym}

\usepackage{booktabs}
\usepackage{amsmath}
\usepackage{threeparttable}
\usepackage{graphicx}
\usepackage{caption}
\usepackage{float}
\usepackage{multirow}

\usepackage{microtype}

\aclfinalcopy 

\title{CAiRE in DialDoc21: Data Augmentation for Information-Seeking Dialogue System}
\author{Etsuko Ishii\thanks{$^*$ These two authors contributed equally.} , Yan Xu$^*$, Genta Indra Winata, Zhaojiang Lin, \\ {\bf Andrea Madotto, Zihan Liu, Peng Xu and Pascale Fung} \\
Center for Artificial Intelligence Research (CAiRE)\\
The Hong Kong University of Science and Technology, Clear Water Bay, Hong Kong\\
\tt \{eishii,yxucb,giwinata\}@connect.ust.hk, pascale@ece.ust.hk
  }

\date{}

\begin{document}
\maketitle
\begin{abstract}
Information-seeking dialogue systems, including knowledge identification and response generation, aim to respond to users with fluent, coherent, and informative responses based on users' needs, which. To tackle this challenge, we utilize data augmentation methods and several training techniques with the pre-trained language models to learn a general pattern of the task and thus achieve promising performance. In DialDoc21 competition, our system achieved 74.95 F1 score and 60.74 Exact Match score in subtask 1, and 37.72 SacreBLEU score in subtask 2. Empirical analysis is provided to explain the effectiveness of our approaches.

\end{abstract}


\section{Introduction}
Recent progress in research has opened up real-life applications of dialogue systems~\cite{winata2021nora,ishii2021erica}, of which information-seeking dialogue systems are one of the major types. The goal of such dialogue systems is to provide fluent and coherent responses with sufficient information to users based on their needs, retrieving information using the dialogue history. The performance of an information-seeking dialogue system can be evaluated from three aspects: (1) user utterance understanding, (2) relevant knowledge retrieval, and (3) agent response generation~\citep{feng2020doc2dial}.

This paper presents work on the DialDoc-21 Shared Task, which is to teach a dialogue system to identify the most relevant knowledge in the associated document for generating agent responses in natural language. It is composed of two subtasks: Knowledge Identification (KI) to retrieve the knowledge from the document, and Response Generation (RG) to generate an agent utterance utilizing the retrieved knowledge.

To tackle this problem, we leverage the pre-trained language models from \citet{liu2019roberta} and \citet{lewis2020bart} and explore data augmentation methods with several training techniques so as to avoid over-fitting to the DialDoc datasets and to teach the model the general pattern of the task. Ensemble and post-processing are conducted to further improve the model performance.
Experimental results show that data augmentation is a simple but effective approach for knowledge identification in information-seeking dialogue systems~\cite{madotto2020learning}, while bringing improvement to response generation at the same time. 
In the DialDoc-21 competition, our system achieved 74.95 of F1 score and 60.74 of Exact Match in subtask 1, and 37.72 SacreBLEU score~\citep{post2018call} in subtask 2\footnote{The code is available at: \url{https://github.com/HLTCHKUST/CAiRE_in_DialDoc21}.}.

\section{Datasets}
\label{sec:data}
\paragraph{Doc2Dial dataset} In this shared task, we mainly focus on the Doc2Dial dataset~\citep{feng2020doc2dial}. Doc2Dial addresses the challenge of modeling different dialogue scenes with documents and providing free-form responses while allowing follow-up questions from the agent. The shared task evaluation is divided into a testdev phase and a test phase. The main difference between these is that in the test phase, out-of-domain (OOD) data samples are included by selecting documents from the domain which is unseen in the training process. The testdev phase only covers 30\% of the data samples in the final test phase. 

Besides Doc2Dial, several other datasets are leveraged for augmentation, as follows:
\paragraph{MRQA 2019 Shared Task dataset} is a collection of multiple reading comprehension datasets for evaluating the generalization ability of QA models. Six datasets are assigned to the training split, which is not included in the evaluation. Among them, SearchQA~\cite{dunn2017searchqa} and TriviaQA~\cite{joshi2017triviaqa} differ from the others by the data resource and have the least generalization ability compared to the other four datasets as reported in~\cite{su2019generalizing}. In this shared task, we consider two settings when leveraging the MRQA dataset: MRQA and MRQA$_{\mathrm{small}}$ which excludes SearchQA and TriviaQA.

\paragraph{Conversational QA (CQA) datasets}
We also introduce three CQA datasets, CoQA~\cite{reddy2019coqa}, QuAC~\cite{choi2018quac}, and DoQA~\cite{campos2020doqa}, in the shared task because of their similar settings to the KI process. 

\paragraph{Wizard-of-Wikipedia (WoW)} is a commonly-used knowledge-grounded dialogue dataset~\cite{dinan2018wizard}. It aims at providing content-full responses to user utterances based on Wikipedia documents. 

\begin{table}[!t]
\centering
\resizebox{\linewidth}{!}{%
\begin{tabular}{@{}llcc@{}}
\toprule
\multirow{2}{*}{Model} & \multirow{2}{*}{Initialization} & \multicolumn{2}{c}{Training} \\ \cmidrule(l){3-4} 
 &  & Data & Method \\ \midrule
\texttt{RoBERTa}$_\mathrm{mrqa}$ & \texttt{RoBERTa}$_\mathrm{large}$ & MRQA & PT \\
\texttt{RoBERTa}$_\mathrm{mrqa_{s}}$ & \texttt{RoBERTa}$_\mathrm{large}$ & MRQA$_{small}$ & PT \\
\texttt{RoBERTa}$_\mathrm{cqa}$ & \texttt{RoBERTa}$_\mathrm{large}$ & Doc2Dial, CQA & FT \\
\texttt{RoBERTa}$_\mathrm{f(cqa)}$ & \texttt{RoBERTa}$_\mathrm{cqa}$ & Doc2Dial & FT \\
\texttt{RoBERTa}$_\mathrm{f(mrqa)}$ & \texttt{RoBERTa}$_\mathrm{mrqa}$ & Doc2Dial & FT \\
\texttt{RoBERTa}$_\mathrm{cqa(mrqa)}$ & \texttt{RoBERTa}$_\mathrm{mrqa}$ &  Doc2Dial, CQA & FT \\
\texttt{RoBERTa}$_\mathrm{cqa(mrqa_s)}$ & \texttt{RoBERTa}$_\mathrm{mrqa_{s}}$ &  Doc2Dial, CQA & FT \\
\texttt{RoBERTa}$_\mathrm{f(cqa(mrqa_s))}$ & \texttt{RoBERTa}$_\mathrm{cqa(mrqa_s)}$ & Doc2Dial & FT \\
\texttt{RoBERTa}$_\mathrm{all}$ & \texttt{RoBERTa}$_\mathrm{large}$ & \begin{tabular}[c]{@{}c@{}}Doc2Dial, CQA, \\ and MRQA\end{tabular} & FT \\ \bottomrule
\end{tabular}
}
\caption{The combinations of the experimental settings for the KI subtask. Two-stage training consists of two stages: pre-training (PT) and fine-tuning (FT).\label{tab:task1-data}}
\end{table}

\section{Methodology}
We utilize a series of data-augmentation approaches to enable the model to obtain better representations on both dialogue context and document context and learn a general pattern of the task with less domain bias.
Namely, we have a two-stage training paradigm, the first step is pretraining (PT) to have a better model initialization, and the second step is fine-tuning (FT) to adapt to DialDoc task. 
For each step, we can apply the multi-task learning (MTL) strategy if we have multiple datasets by making the datasets format uniform and treat samples equally.
As reported in~\citet{fisch2019mrqa}, a model trained on multiple dataset under similar tasks, is supposed to provide a better initialization for further fine-tuning and is capable of generalizing to the data samples in other domains. 
Thus, we expect a model trained with MTL in the first step to offer a better initialization and in the second step to reduce the domain bias and avoid overfitting.

\subsection{Knowledge Identification}
\label{subsec:task1}
In the KI task, we conduct experiments on a large pre-trained model, RoBERTa-large~\cite{liu2019roberta}, which has shown its effectiveness on many QA datasets~\cite{ju2019technical}. The MRQA dataset and three CQA above datasets are leveraged for data augmentation. The combinations of the experimental settings are considered as follows:

We consider using CQA datasets to enrich the data source. \textbf{\texttt{RoBERTa}$_\mathrm{cqa}$} is fine-tuned on Doc2Dial and three CQA datasets using MTL method. \textbf{\texttt{RoBERTa}$_\mathrm{f(cqa)}$} leverages the pre-trained \texttt{RoBERTa}$_\mathrm{cqa}$ model and is fine-tuned on Doc2Dial dataset for better performance.

We train the RoBERTa model on MRQA daztaset and MRQA$_\mathrm{small}$ dataset described in \S~\ref{sec:data} using MTL respectively (denoted as \texttt{RoBERTa}$_\mathrm{mrqa}$ and \texttt{RoBERTa}$_\mathrm{mrqa_{s}}$). These models could be further fine-tuned while providing a better initialization~\cite{fisch2019mrqa}. \textbf{\texttt{RoBERTa}$_\mathrm{f(mrqa)}$} is to further fine-tune \texttt{RoBERTa}$_\mathrm{mrqa}$ on Doc2Dial dataset. The corresponding settings are also applied to \textbf{\texttt{RoBERTa}$_\mathrm{f(mrqa_s)}$} model. While \textbf{\texttt{RoBERTa}$_\mathrm{cqa(mrqa)}$} is initialized with \texttt{RoBERTa}$_\mathrm{mrqa}$ and fine-tuned on Doc2Dial and three CQA datasets using MTL. \textbf{\texttt{RoBERTa}$_\mathrm{cqa(mrqa_s)}$} follows the same setting as the former model, but use \texttt{RoBERTa}$_\mathrm{mrqa_s}$ model for initialization instead. \textbf{\texttt{RoBERTa}$_\mathrm{f(cqa(mrqa_s))}$} is to further fine-tune \texttt{RoBERTa}$_\mathrm{cqa(mrqa_s)}$ on Doc2Dial dataset.

\textbf{\texttt{RoBERTa}$_\mathrm{all}$} is trained on Doc2Dial, MRQA dataset and CQA datasets using MTL method.

For better readability, we summarize the model settings in Table~\ref{tab:task1-data}. We also explore more combinations of the experimental settings, such as other combinations of the datasets and other pre-trained language models. However, those fail to bring the improvements as much as those we mentioned above.

\paragraph{Post-processing} We further conduct post-processing on the model predictions based on our observation that the ground truths of the data samples are annotated by document splits which are provided together with the dataset. We consider including the whole split of the document once the prediction covers $\lambda$ percent of it, where $\lambda$ is set as 0.1. In addition, for better performance in the shared task, we also slightly extend the predictions when there is a ``Yes'' or ``No'' shown right in front of the predicted spans.

\paragraph{Ensemble} To further boost the model performance, we build an ensemble of our existing models. We consider one prediction containing the start position and the end position of the document as a unit and conduct voting over all the predictions of each data sample. The most frequent one will be selected as the final prediction. We denote the ensemble result as \texttt{RoBERTa}$_\mathrm{ensemble}$.

\begin{table}[!th]
\centering
\resizebox{\linewidth}{!}{%
\begin{tabular}{@{}ll|ll@{}}
\toprule
\multicolumn{2}{l|}{Knowledge Identification} & \multicolumn{2}{l}{Response Generation} \\ \midrule
max input length & 512 & max input length & 300 \\
max answer length & 50 & max target length & 200 \\
batch size & 120 & batch size & 60 \\
document stride & 128 & beam size & 4 \\
learning rate & 3e-5 & learning rate & 3e-5 \\\bottomrule
\end{tabular}
}
\caption{The hyper-parameter settings in the shared task.\label{tab:hyper}}
\end{table}

\subsection{Response Generation}
To obtain natural and relevant responses, we take advantage of the evidence to the query identified from \S~\ref{subsec:task1} and focusing on paraphrasing the corresponding knowledge sentences based on the dialogue context. We leverage the large pre-trained model BART$_{\mathrm{large}}$~\cite{lewis2020bart}. The process of training and inference can be summarized as three steps:
\paragraph{Pre-training on WoW dataset.} We first pre-train the BART model on the WoW dataset for better initialization because of its similarity with the RG task. In the training process, the gold grounded knowledge sentences are concatenated with the dialogue context and fed into the model as the inputs.

\paragraph{Fine-tuning on Doc2Dial dataset.} In the Doc2Dial dataset, the labels of the gold document splits are also provided in the training and validation set. The model is further fine-tuned on the Doc2Dial dataset using the same components for the input sequences in the first step. The model could be evaluated under two scenarios: (1) \textbf{Gold mode} (\texttt{BART}$_\mathrm{gold}$), leveraging the gold labels of the knowledge evidence in the dataset as the knowledge inputs; (2) \textbf{Prediction mode} (\texttt{BART}$_\mathrm{pred}$), leveraging the prediction of the KI process as the inputs.

\paragraph{Inference with Knowledge Evidence.} During the testdev and test phase, we leverage the predictions from the KI process as the knowledge evidence components for the dialogue queries. The model generates responses based on a concatenation of the knowledge evidence and the dialogue context.

\paragraph{Post-processing} To avoid serious information loss in the generations compared to the knowledge evidence for the OOD data samples, we compare the lengths of the knowledge evidence and the responses (denoted as $L_\mathrm{kn}$ and $L_\mathrm{resp}$). The generated response will be replaced by the raw knowledge evidence as the final output if $L_\mathrm{resp}\leq\alpha L_\mathrm{kn}$, where $\alpha$ is set as 0.4.


\begin{table}[t!]
\centering
\resizebox{\linewidth}{!}{%
\begin{tabular}{@{}lccc@{}}
\toprule
Model                      & \# of ckpt & EM & F1 \\ \midrule \midrule
Testdev Phase & & & \\ \midrule 
\texttt{RoBERTa}$_\mathrm{large}$ (baseline)   & -      & 58.08   & 72.17   \\ \midrule
\texttt{RoBERTa}$_\mathrm{cqa}$            & 2          & 59.09($\pm$1.01)   & 72.90($\pm$0.25)   \\
\texttt{RoBERTa}$_\mathrm{f(cqa)}$         & 2          & 58.08($\pm$1.01)   & 72.23($\pm$0.18)   \\
\texttt{RoBERTa}$_\mathrm{f(mrqa)}$        & 1          & 58.08   & 72.30   \\
\texttt{RoBERTa}$_\mathrm{f(mrqa_s)}$      & 16         & 59.37($\pm$1.89)   & 73.51($\pm$1.60)   \\
\texttt{RoBERTa}$_\mathrm{cqa(mrqa)}$      & 1          & 58.08   & 72.59   \\
\texttt{RoBERTa}$_\mathrm{cqa(mrqa_s)}$    & 6          & 59.60($\pm$1.35)   & 73.76($\pm$1.57)   \\
\texttt{RoBERTa}$_\mathrm{f(cqa(mrqa_s))}$ & 1          & 60.10   & 75.02   \\
\texttt{RoBERTa}$_\mathrm{all}$            & 1          & 58.08   & 74.63   \\ 
\texttt{RoBERTa}$_\mathrm{ensemble}$       & -          & 63.13   & 77.31   \\ \midrule
\texttt{RoBERTa}$_\mathrm{all}$-postproc       & -      & 57.07(-1.01)   & 74.15(-0.47)   \\
\texttt{RoBERTa}$_\mathrm{ensemble}$-postproc  & -      & 63.13(-0.00)  & 76.73(-0.58)   \\ \midrule \midrule
Test Phase & & & \\ \midrule 
\texttt{RoBERTa}$_\mathrm{ensemble}^*$  & -      & 60.74  & 74.95   \\ \bottomrule
\end{tabular}
}
\caption{The results of the selected models on the testdev and test phase of subtask 1 are listed. All the results are calculated with the corresponding predictions after post-processing except those with specific notations. For the models that are trained with multiple random seeds, the average scores and the standard deviations are presented. \texttt{RoBERTa}$_\mathrm{ensemble}^*$ denotes the results of the ensemble model on the test set.  \label{tab:task1}} 
\end{table}

\section{Experiments}
\subsection{Training Details}
\paragraph{Hyper-parameter Settings} We apply different settings to utilize the dialogue history for the two subtasks. For subtask 1, we leverage all previous turns and build the input sequence in a reverse order to them. For subtask 2, we leverage one extra last turn in the time order and differentiate the speakers with special tokens. In Table~\ref{tab:hyper}, we list the selected hyper-parameters utilized in the shared task. 


\paragraph{Ensemble Settings} In subtask 1, we make an ensemble of all the checkpoints of the models listed in Table~\ref{tab:task1-data} except \texttt{RoBERTa}$_\mathrm{mrqa}$ and \texttt{RoBERTa}$_\mathrm{mrqa_{s}}$. The details of the checkpoints can be found in Tabel~\ref{tab:task1}. 

\paragraph{Metrics and Model Selection} 
In subtask 1, the Exact Match (EM) and uni-gram F1 score are utilized as the criteria, while in subtask 2, we evaluate the generation by SacreBLEU. We select the models with the best EM and SacreBLEU scores on the validation set respectively, for the two subtasks. Specifically for subtask 2, the model is selected under the gold mode.

\begin{table}[t]
\centering
\resizebox{\linewidth}{!}{%
\begin{tabular}{@{}lccc@{}}
\toprule
\multirow{2}{*}{Model} & \multicolumn{3}{c}{SacreBLEU} \\ \cmidrule(l){2-4} 
 & val & testdev & test \\ \midrule
\texttt{BART}$_\mathrm{large}$ (baseline)  & -      & 16.73  & - \\
\texttt{Gold}                              & 45.67  & -      & - \\
\texttt{RoBERTa}$_\mathrm{ensemble}$       & 38.78  & 37.45  & 38.68 \\ \midrule
\texttt{BART}$_\mathrm{gold}$              & 20.17  & -      & - \\
\quad +WoW pre-traning                     & 48.24  & -      & - \\ 
\texttt{BART}$_\mathrm{pred}$              & 16.67  & 16.72  & 16.45  \\
\quad +WoW pre-traning                     & 39.87  & 38.26  & 37.31  \\
\quad +WoW pre-training+postproc$^*$       & -      & -      & 37.72  \\ \bottomrule
\end{tabular}
}
\caption{The results of selected models on subtask 2 are listed. \texttt{Gold} denotes the gold knowledge evidence labels provided in the dataset. The model denoted with $^*$ is the final submission to the test phase. \label{tab:task2}} 
\end{table}

\subsection{Results and Discussion}
\subsubsection{Results} 
The results are shown in Table~\ref{tab:task1} and Table~\ref{tab:task2}. For both subtasks, we observe gaps between the testdev phase and the test phase. For some of the models in subtask 1, multiple random seeds are applied in the training process. The performance gap may result from the domain difference of the partial data samples in the test phase, where the corresponding documents are unseen in the training set. In Table~\ref{tab:task1}, without post-processing on the predictions, the model performance consistently drops to a certain extent, which indicates that post-processing is suitable for the Doc2Dial scenario. Ensemble, which is a common strategy to improve performance, shows its effectiveness in this task. 

For subtask 2, the pre-training on WoW dataset brings huge improvement to the model. Interestingly, by just using the knowledge evidence predicted from the subtask 1 \texttt{RoBERTa}$_\mathrm{ensemble}$ model or the gold knowledge evidence labels, the performance can even exceed that of the generative model on SacreBLEU scores, while the responses from \texttt{BART}$_\mathrm{pred}$ are more fluent and natural. This may be caused by the information loss when paraphrasing the knowledge evidence to dialogue responses.


\subsubsection{Discussion}
In this task, we explore data augmentation methods and conduct two-stage training as auxiliary training strategy for improvement. Although resource- and time-consuming, this approach is easy to implement and effective at enabling the model to learn more general ability on the task. 

\subsubsection{Post-Challenge Improvements}
From our findings, the hyper-parameter, the maximum answer length, is left untuned, which hurts the QA model performance to some degree. With a maximum answer length of 100, the EM and F1 score on the testdev set improve by 2.53 and 1.08, respectively, while a 64.42 EM and 77.27 F1 score are achieved on the test set. With the improved prediction from subtask 1, we achieve a 39.88 SacreBLEU score in subtask 2.

\section{Related Work}
Conversational QA is a type of reading comprehension task that requires understanding not only the question but also the previous conversation turns. Various datasets have been introduced in recent years, and many of them restrict answers to be extraction of a span from the reference document, while the others allow free-form responses~\citep{choi2018quac, reddy2019coqa, campos2020doqa}. 

In addition to the works to enrich the contents of open-domain conversations by controllable generation~\citep{lin2020xpersona, madotto2020plug}, the knowledge grounded dialogue task aims to offer more informative conversation by leveraging an external knowledge source~\citep{dinan2018wizard,xu2020controllable}. Relevant knowledge selection is the key to improving the whole system, and very recently, latent variable models have been attracting more attention for this purpose~\citep{lian2019learning, liu2019zero, Kim2020Sequential, chen2020bridging, xu2021retrieval}.

\section{Conclusion}
In this paper, we utilize data augmentation methods and several training techniques with pre-trained language models to tackle the challenge of the information-seeking dialogue task. The results have indicated the effectiveness of our approaches. Moreover, data augmentation methods are easy to implement, which is promising for practical use.

\bibliographystyle{acl_natbib}
\bibliography{acl2021}


\end{document}